\documentclass[conference]{IEEEtran}
\IEEEoverridecommandlockouts
\usepackage{cite}
\usepackage{amsmath,amssymb,amsfonts}
\usepackage{graphicx}
\usepackage{textcomp}

\usepackage[colorlinks=true, linkcolor=blue, urlcolor=blue, citecolor=blue]{hyperref}

\begin{document}

\title{BengaliSent140: A Large-Scale Bengali Binary Sentiment Dataset for Hate and Non-Hate Speech Classification}

\author{
Akif Islam\textsuperscript{1},
Sujan Kumar Roy\textsuperscript{1},
Md.~Ekramul Hamid\textsuperscript{1}
\\[1ex]
\textsuperscript{1}Department of Computer Science and Engineering, University of Rajshahi, Bangladesh

}

\maketitle

\begin{abstract}
Sentiment analysis for the Bengali language has gained notable research interest in recent years; however, meaningful progress is often limited by the lack of sufficiently large and diverse annotated datasets. Although several Bengali sentiment and hate speech datasets are publicly available, most of them are either small in scale or collected from a single domain, such as social media comments. As a result, these datasets are often inadequate for training modern deep learning–based models, which require large volumes of heterogeneous data to learn robust and generalizable representations. In this work, we introduce \textbf{BengaliSent140}, a large-scale Bengali binary sentiment dataset created by consolidating seven existing Bengali text datasets into a unified corpus. To ensure consistency, heterogeneous annotation schemes are harmonized into a binary sentiment formulation with two classes: \emph{Not Hate (0)} and \emph{Hate (1)}. The final dataset contains \textbf{139,792 unique text samples}, including \textbf{68,548 hate} and \textbf{71,244 not-hate} instances, resulting in a relatively balanced class distribution. By merging data from multiple sources and domains, BengaliSent140 offers broader coverage than existing Bengali sentiment datasets and provides a suitable foundation for training and benchmarking deep learning models. Baseline experimental results are also reported to demonstrate the practical usability of the dataset. The dataset is publicly available at \url{https://www.kaggle.com/datasets/akifislam/bengalisent140}.
\end{abstract}


\section{Introduction}

The way people communicate has changed dramatically with the rise of digital platforms. Today, opinions, emotions, and personal experiences are constantly shared through social media, online news portals, and video-sharing services. Understanding these expressions at scale has made sentiment analysis a central problem in Natural Language Processing (NLP). It now underpins a wide range of real-world applications, from public opinion analysis and content moderation to social media monitoring and decision support systems \cite{liu2012sentiment}. As online content continues to grow at an unprecedented rate, automated sentiment analysis has become essential for making sense of large-scale social behavior.

Bengali is one of the most widely spoken languages in the world, used by more than 230 million people, primarily in Bangladesh and India. Over the past decade, the rapid expansion of digital platforms has significantly increased the volume of Bengali text available online \cite{sarkar2024socialmediagrowth}. These platforms have opened new spaces for public participation, allowing people to express opinions, debate political and social issues, and engage with news in their native language. However, alongside constructive discussion, these spaces have increasingly become channels for negative sentiment, hate speech, and abusive language.

In Bangladesh, online platforms are often misused for harmful activities such as political and religious intimidation, coordinated harassment, and gender-based abuse. Women—particularly adolescent women—are disproportionately targeted in many of these interactions \cite{islam2025parameterefficientfinetuninglowresourcelanguages}. The growing visibility of such hostility highlights an urgent need for reliable sentiment analysis and hate speech detection systems that can effectively operate in the Bengali language.

Political discourse further intensifies this challenge. Elections, political movements, protests, and policy debates frequently spark emotionally charged and highly polarized discussions online, where aggressive language and hate-driven narratives are common \cite{mathew2019hate}. Television talk shows and political debate programs often amplify these dynamics through confrontational rhetoric that shapes public sentiment beyond traditional media boundaries. Automated analysis of this content can offer valuable insights into public mood, ideological polarization, and early indicators of social or political unrest.

Beyond politics, online harassment, bullying, and cyberbullying are widespread across Bengali digital spaces. Comment sections on social media and news platforms regularly contain abusive language targeting gender, religion, ethnicity, or political identity. Given the scale, speed, and anonymity of these platforms, manual moderation is no longer feasible, making automated sentiment and hate speech detection systems a practical necessity \cite{vidgen2019challenges}. Robust sentiment analysis models can play an important role in identifying harmful content, supporting moderation efforts, and promoting safer online environments.

Despite the clear societal importance of sentiment analysis for Bengali, progress in this area remains limited by the lack of large-scale, high-quality annotated datasets. Most existing Bengali sentiment or hate speech datasets are either small, confined to specific domains such as social media comments, or annotated using inconsistent labeling schemes such as toxic, abusive, offensive, or cyberbullying. These constraints pose serious challenges for training modern deep learning models, which typically require large, diverse, and well-labeled data to generalize effectively \cite{young2018recent}.

Motivated by this gap, we introduce \textbf{BengaliSent140}, a large-scale Bengali binary sentiment dataset created to support robust hate and non-hate speech classification. The dataset is constructed by consolidating seven publicly available Bengali text corpora and systematically harmonizing their diverse annotation schemes into a unified binary framework. The resulting corpus contains approximately \textbf{140K unique text samples}, from which the name \textit{BengaliSent140} is derived, including \textbf{68,548 hate} and \textbf{71,244 not-hate} instances. This scale and balanced class distribution make the dataset particularly suitable for training deep learning–based sentiment analysis models. By integrating data from multiple sources and domains, BengaliSent140 offers broader linguistic and contextual coverage than existing Bengali datasets and supports more generalized and reliable sentiment modeling.

The main contributions of this work are summarized as follows:
\begin{enumerate}
    \item We construct a large-scale Bengali binary sentiment dataset by consolidating seven heterogeneous Bengali text corpora into a unified benchmark.
    \item We harmonize diverse and inconsistent annotation schemes into a consistent hate versus not-hate classification formulation.
    \item We release multiple textual representations, including raw, normalized, and lemmatized versions, to facilitate systematic analysis of preprocessing effects.
    \item We provide baseline experimental results using classical machine learning and deep learning models to establish reference benchmarks.
\end{enumerate}

\section{Related Work}

In recent years, several datasets have been introduced to support sentiment analysis, abusive language detection, and hate speech classification in Bengali. One of the largest early efforts is the BD-SHSH dataset \cite{romim2022bdshs}, which contains approximately 49,995 text samples primarily collected from social media platforms. While BD-SHSH provides a relatively large corpus, its annotation scheme is not strictly binary and includes heterogeneous labels related to hate and abuse. Moreover, its strong focus on social media comments limits its applicability across other important domains such as political discourse or news media.

The Dataset for Cyberbully Detection in Bengali Comments \cite{ahmed2021cyberbullyingdetectionusingdeep} consists of around 43,567 text instances and is designed specifically to identify cyberbullying behavior. Although the dataset is sizable, it is narrowly focused on bullying-related phenomena rather than general sentiment or hate speech classification. As a result, its labeling strategy and task formulation are not directly aligned with binary sentiment analysis or broader hate speech detection tasks.

SentNoB \cite{islam-etal-2021-sentnob-dataset} is a well-known dataset for sentiment analysis in noisy Bengali texts, containing 14,856 samples annotated for sentiment polarity. Unlike hate speech datasets, SentNoB is primarily formulated as a multi-class sentiment classification problem, typically involving positive, negative, and neutral labels. While valuable for studying sentiment in informal and noisy text, the dataset does not explicitly address hate or abusive language and remains limited in scale for training deep learning models effectively.

The Multi-Labeled Bengali Toxic Comments dataset \cite{multitoxic} includes approximately 11,890 samples annotated with multiple toxicity-related labels such as toxic, obscene, or threatening. Although this multi-label design allows fine-grained analysis, it introduces additional complexity and label sparsity. Furthermore, the dataset size is relatively small, and its annotation scheme is not directly compatible with binary sentiment classification without substantial preprocessing or label consolidation.

Another notable resource is the Bengali Hate Speech Dataset by Romim et al.\ \cite{romimhate}, which contains 9,471 samples explicitly annotated for hate speech detection. While the dataset provides clear hate-related annotations, its limited size and narrow domain coverage restrict its usefulness for training modern deep learning architectures, which generally require larger and more diverse datasets to generalize well.

The Bengali Hate Speech Dataset proposed by Karim \cite{karim2021deephateexplainer} further refines hate speech annotation but contains only 5,604 text samples. Despite careful curation, the small scale of this dataset poses significant challenges for deep learning-based sentiment analysis, often resulting in overfitting and unstable performance. Similarly, the Multimodal-Hate-Bengali dataset \cite{karim2022multimodalhate}, which includes 4,409 samples combining textual and visual information, is valuable for multimodal research but remains limited in size for text-only sentiment or hate speech classification tasks.

Overall, existing Bengali sentiment and hate speech datasets suffer from one or more critical limitations, including insufficient scale, non-binary or inconsistent labeling schemes, domain specificity, and task misalignment. These issues hinder fair benchmarking and restrict the effective use of deep learning architectures. Motivated by these limitations, we unify seven heterogeneous datasets into \textbf{BengaliSent140}, a large-scale and balanced binary sentiment dataset designed specifically to support robust hate and non-hate speech classification in Bengali.

\section{Dataset Construction} 

\begin{table*}[!t]
\centering
\caption{Sample entries from the BengaliSent140 dataset showing Bengali text instances and their corresponding binary labels, where 0 denotes \emph{Not Hate} and 1 denotes \emph{Hate}.}
\label{tab:dataset_samples}
\includegraphics[width=0.7\textwidth]{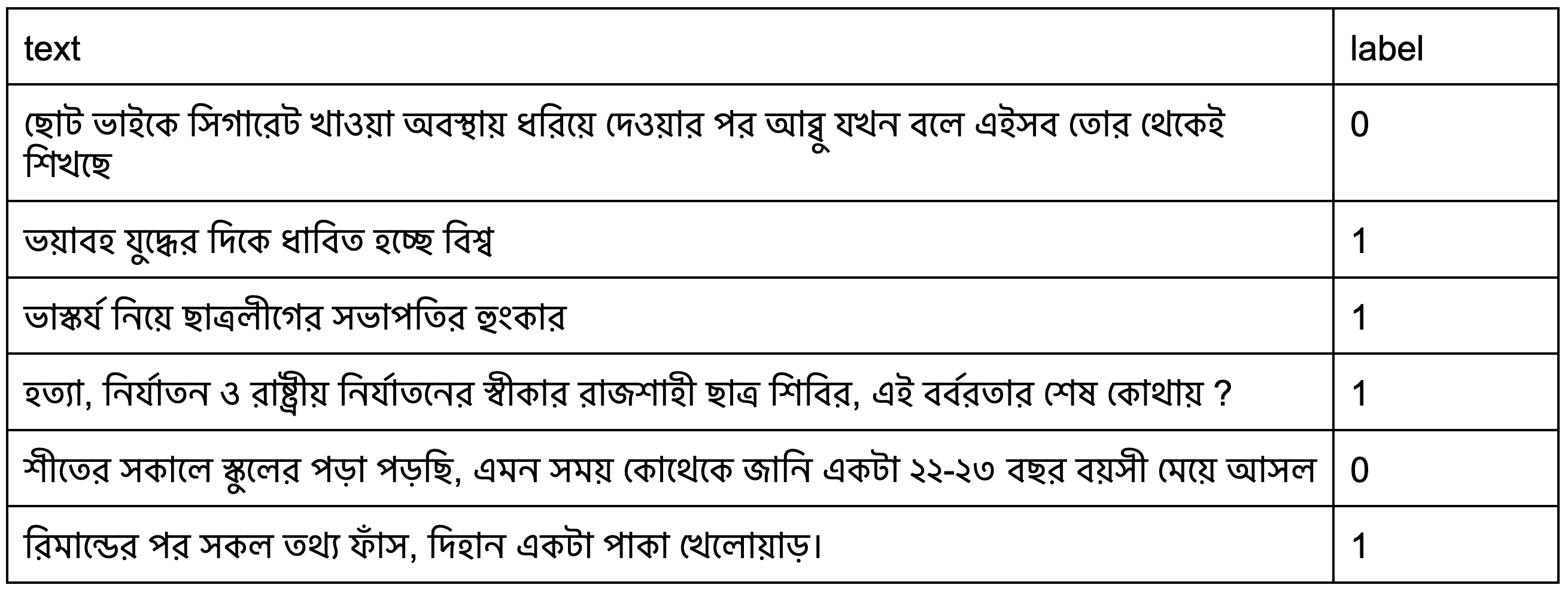}
\end{table*}

\subsection{Source Datasets}

\begin{figure*}[t]
\centering
\includegraphics[width=0.8\textwidth]{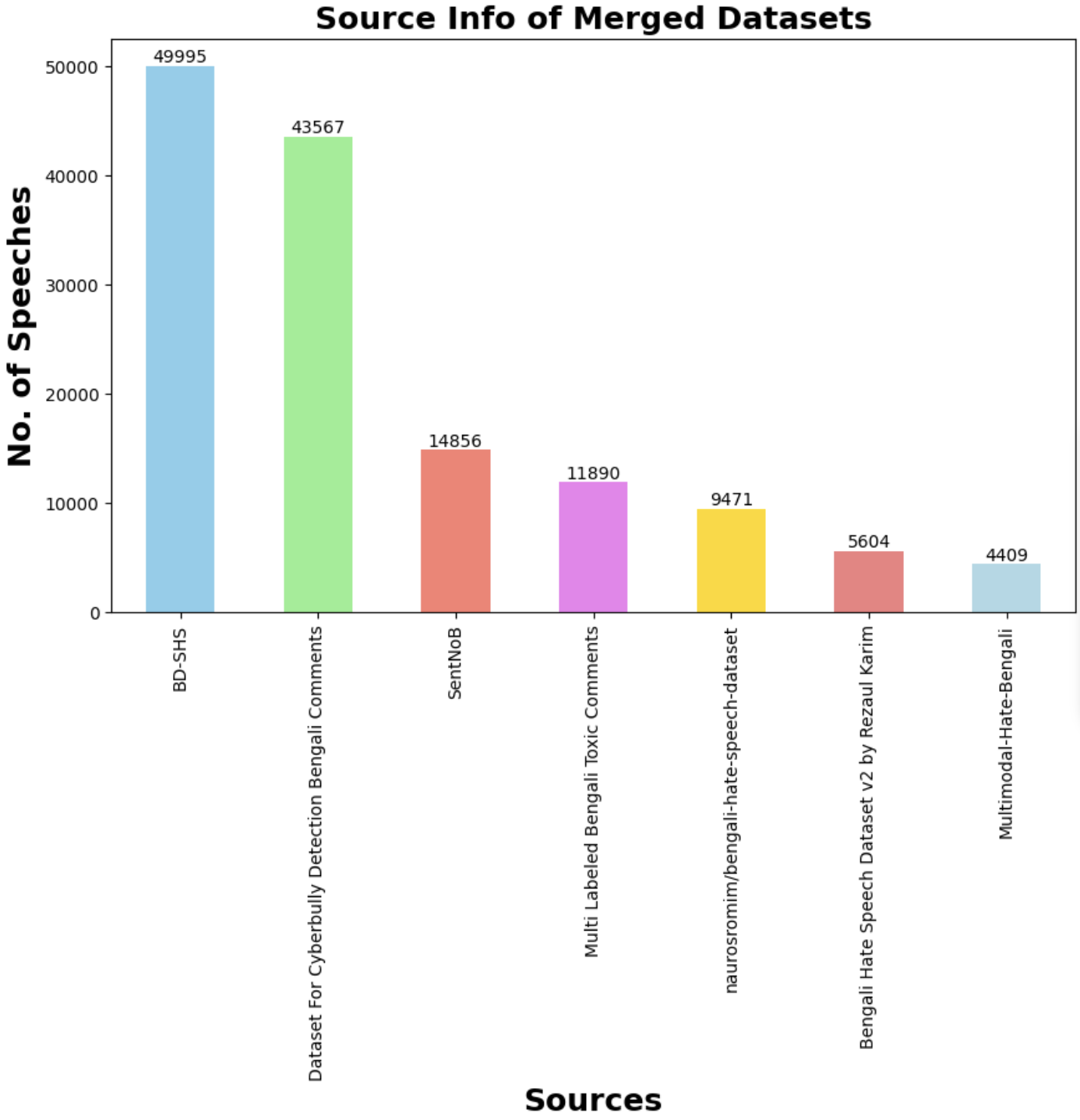}
\caption{Source-wise contribution of text samples from the seven datasets merged to construct the BengaliSent140 corpus, illustrating the relative size and diversity of each data source.}
\label{fig:source_datasets}
\end{figure*}

BengaliSent140 is constructed by merging seven publicly available Bengali text datasets sourced from Kaggle and GitHub repositories and systematically re-labeling them into a unified binary sentiment classification framework. Heterogeneous annotation schemes across the source datasets are harmonized into two classes, namely \emph{Not Hate} and \emph{Hate}, to ensure label consistency. Table~\ref{tab:source_datasets} summarizes the contribution of each individual dataset. The complete BengaliSent140 dataset is publicly released on Kaggle to support open research and reproducibility and is available at \url{https://www.kaggle.com/datasets/akifislam/bengalisent140/}.

\begin{table}[!t]
\caption{Source Datasets Used in BengaliSent140}
\label{tab:source_datasets}
\centering
\begin{tabular}{l c}
\hline
\textbf{Name of Source Dataset} & \textbf{No. of Unique Data} \\
\hline
BD-SHSH \cite{romim2022bdshs} & 49,995 \\
Cyberbully Detection Bengali Comments \cite{ahmed2021cyberbullyingdetectionusingdeep} & 43,567 \\
SentNoB \cite{islam-etal-2021-sentnob-dataset} & 14,856 \\
Multi-Labeled Bengali Toxic Comments \cite{multitoxic} & 11,890 \\
Bengali Hate Speech Dataset \cite{rezav1dataset} & 9,471 \\
Bengali Hate Speech Dataset v2 \cite{karim2021deephateexplainer} & 5,604 \\
Multimodal-Hate-Bengali \cite{karim2022multimodalhate} & 4,409 \\
\hline
\textbf{Total} & \textbf{139,792} \\
\hline
\end{tabular}
\end{table}

The original source datasets employ diverse labeling schemes, including hate, abusive, toxic, cyberbullying, and neutral categories. To ensure annotation consistency across heterogeneous sources, all labels indicating any form of hate, abuse, or harmful intent are unified and mapped to a single class, \emph{Hate (1)}. Conversely, labels corresponding to neutral, non-hateful, or benign content are mapped to \emph{Not Hate (0)}. This label harmonization process enables the formulation of a consistent binary sentiment classification task while preserving the semantic intent of the original annotations.

\begin{figure}[!t]
\centering
\includegraphics[width=0.8\linewidth]{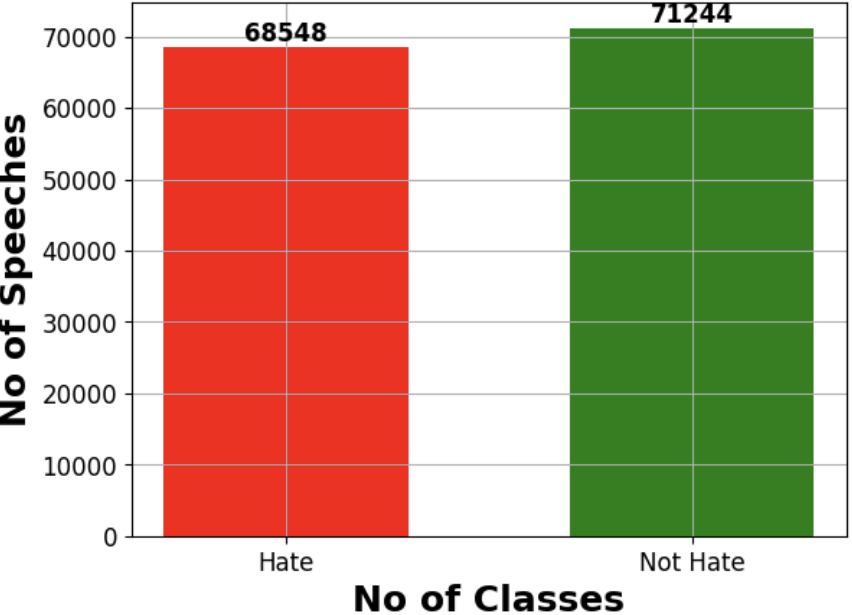}
\caption{Binary class distribution of the BengaliSent140 dataset, illustrating the number of hate and not-hate text samples.}
\label{fig:class_distribution}
\end{figure}

Following label harmonization, the final merged dataset exhibits a relatively balanced class distribution, consisting of 68,548 hate samples and 71,244 not-hate samples, as illustrated in Fig.~\ref{fig:class_distribution}. Such balance is particularly important for sentiment and hate speech classification, as it helps reduce classifier bias toward majority classes and contributes to more stable and reliable model training and evaluation. Nevertheless, the classification task remains inherently challenging. Even state-of-the-art deep learning and large language models often struggle to reliably distinguish between hate and not-hate content when expressions are implicit, context-dependent, or convey double or subtle meanings.

\subsection{Data Preprocessing}

\begin{table}[!t]
\centering
\caption{Illustration of Bengali text preprocessing showing original, normalized (punctuation removed), and lemmatized representations.}
\label{tab:preprocessing_example}
\includegraphics[width=0.85\linewidth]{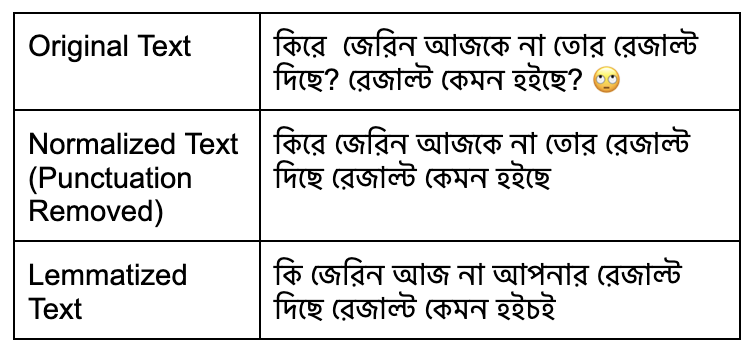}
\end{table}

Figure~\ref{tab:preprocessing_example} illustrates the basic preprocessing steps applied to the BengaliSent140 dataset. To enable systematic analysis of preprocessing effects, we preserve multiple textual representations for each sample rather than enforcing a single normalized form. This design allows researchers to directly compare model behavior across different input variants without reprocessing the raw data.

Each sample contains four fields: \textit{text}, \textit{no\_punc\_text}, \textit{lemmatized\_text}, and \textit{label}. The \textit{text} field stores the original unprocessed Bengali text, which may include emojis, punctuation, and informal expressions. The \textit{no\_punc\_text} field removes emojis, punctuation, and hashtags to reduce surface-level noise, while \textit{lemmatized\_text} applies lemmatization to the normalized text. Due to limitations of existing Bengali lemmatization tools, some words cannot be accurately reduced to their root forms, these artifacts are intentionally retained to reflect realistic preprocessing conditions. The \textit{label} field contains the final binary annotation, where hate-related content is mapped to 1 and non-hate content to 0.

For model training, we explored several word embedding techniques, including Word2Vec, GloVe, FastText, and pre-trained FastText embeddings, to convert text into numerical representations. The vocabulary was limited to the 10,000 most frequently occurring tokens, and sequences were padded to a maximum length of 100. While this setup is effective for conventional neural architectures, it may not be optimal for all scenarios, as newer transformer-based models such as BERT often benefit from preserving punctuation and richer contextual cues.

\section{Experimental Setup}

We conduct a series of baseline experiments using both classical machine learning and deep learning models with a NVIDIA RTX 3060 Ti GPU. The objective of these experiments is to provide reproducible reference results that future studies can use for fair comparison, rather than to pursue state-of-the-art performance.

\subsection{Models}

We evaluate four representative classifiers that capture different modeling paradigms. A Fully Connected Neural Network (FCNN) is used as a simple neural baseline to assess the effectiveness of learned word representations without explicit sequence modeling. A Bidirectional Long Short-Term Memory (Bi-LSTM) network is employed to capture contextual dependencies in both forward and backward directions, which is particularly important for sentiment analysis in morphologically rich languages such as Bengali. To further enhance local feature extraction, a hybrid Convolutional Neural Network and Bi-LSTM (Conv + Bi-LSTM) architecture is evaluated, combining convolutional layers for n-gram feature learning with recurrent layers for long-range dependency modeling. In addition to neural models, a Random Forest (RF) classifier is included as a classical machine learning baseline to provide a point of comparison against deep learning approaches.

\subsection{Embedding Strategies}

Two embedding strategies are explored for text representation. First, trainable Keras word embeddings are learned directly during model training, allowing the embeddings to adapt specifically to the BengaliSent140 corpus. Second, pre-trained FastText embeddings are employed to incorporate subword-level information, which is particularly effective for handling spelling variations and out-of-vocabulary words commonly observed in Bengali text. These two strategies enable comparative analysis between corpus-specific embeddings and externally trained representations.

\subsection{Evaluation Metrics}

Model performance is evaluated using standard classification metrics, including accuracy, loss, and F1-score on a held-out test set. While accuracy provides an overall measure of prediction correctness and loss reflects the optimization behavior during training, F1-score is emphasized as the primary evaluation metric. F1-score offers a balanced evaluation of precision and recall and is especially suitable for binary sentiment and hate speech classification tasks, where both false positives and false negatives have significant practical implications.

\section{Results and Discussion} 
\subsection{Model Performance on BengaliSent140}
\begin{table}[htbp]
\caption{Performance Comparison of Machine Learning, Deep Learning, and Transfer Learning Models on BengaliSent140 dataset}
\label{tab:model_comparison}
\centering
\begin{tabular}{l c}
\hline
\textbf{Model} & \textbf{Test Accuracy} \\
\hline
\multicolumn{2}{l}{\textbf{Machine Learning Models}} \\
Logistic Regression & 0.79 \\
SVM & 0.79 \\
Random Forest & 0.83 \\
XGBoost & 0.84 \\
\hline
\multicolumn{2}{l}{\textbf{Deep Learning Models}} \\
FCNN & 0.81 \\
LSTM & 0.84 \\
CNN & 0.85 \\
Bi-LSTM & 0.87 \\
ConvLSTM & 0.87 \\
Dilated Convolution & 0.88 \\
\hline
\multicolumn{2}{l}{\textbf{Transfer Learning}} \\
BERT & 0.91 \\
\hline
\end{tabular}
\end{table}

Table~\ref{tab:model_comparison} reports a comparative performance analysis of classical machine learning, deep learning, and transfer learning models evaluated on the BengaliSent140 dataset. As expected, traditional machine learning approaches such as Logistic Regression and SVM achieve moderate test accuracies of 0.79, indicating their limited ability to capture complex linguistic patterns in Bengali text. Tree-based methods, including Random Forest and XGBoost, perform comparatively better, reaching accuracies of 0.83 and 0.84 respectively, benefiting from non-linear feature interactions.

Deep learning models consistently outperform classical approaches, demonstrating the advantage of representation learning on large-scale textual data. The FCNN model achieves a test accuracy of 0.81, while sequence-aware architectures such as LSTM and Bi-LSTM further improve performance, reaching 0.84 and 0.87 respectively. Convolution-based models, including CNN and ConvLSTM, also show strong performance, highlighting the effectiveness of local feature extraction for sentiment classification. The Dilated Convolution model achieves the highest accuracy among non-transformer models at 0.88, suggesting its ability to capture longer contextual dependencies.

The transfer learning approach using BERT yields the best overall performance with a test accuracy of 0.91. This result underscores the strength of transformer-based architectures and pre-trained language models in handling contextual and semantic complexity, particularly in low-resource languages like Bengali. Overall, these results demonstrate that BengaliSent140 supports effective benchmarking across a wide range of modeling paradigms and can serve as a reliable evaluation resource for both conventional and advanced sentiment analysis models.

\subsection{Challenges and Limitations}

\begin{figure}[!b]
\centering
\includegraphics[width=0.65\linewidth]{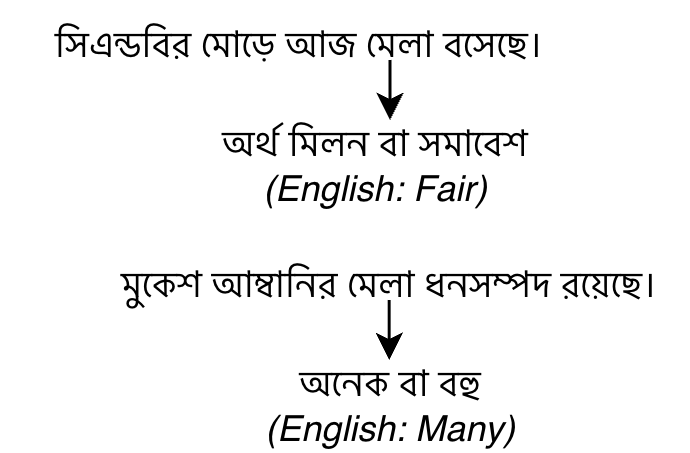}
\caption{Illustration of semantic ambiguity in Bengali, where identical words convey different meanings depending on context, posing challenges for sentiment and hate speech classification.}
\label{fig:semantic_ambiguity}
\end{figure}

Although BengaliSent140 is large in scale and constructed from diverse sources, several challenges and limitations remain. The dataset primarily consists of monolingual Bengali text and does not explicitly capture code-mixed usage of Bengali and English, which is increasingly prevalent among the Gen-Z population in Bangladesh, particularly on social media platforms. Furthermore, the dataset is formulated as a binary classification task with labels restricted to \emph{Hate} and \emph{Not Hate}. While this design is practical for many applications, it does not capture finer-grained distinctions such as political, religious, or personal hate, largely due to the absence of consistently annotated multi-class labels across the source datasets. 

Another significant challenge stems from the inherent semantic ambiguity of the Bengali language, where the same word or phrase may convey different meanings depending on context, as illustrated in Figure~\ref{fig:semantic_ambiguity}. Such ambiguity makes it difficult to distinguish implicit or context-dependent hate, even for modern deep learning models. Additionally, the preprocessing pipeline relies on existing Bengali lemmatization tools with limited accuracy, which may introduce minor normalization noise but reflects realistic preprocessing conditions. Finally, the experimental results presented in this work are based on lightly tuned baseline models, as the primary focus is the introduction and documentation of the dataset - more advanced architectures, extensive optimization, and context-aware large language models are left for future exploration.

\section{Conclusion and Future Work}

This paper presents \textbf{BengaliSent140}, a large-scale Bengali binary sentiment dataset created by merging multiple heterogeneous sources into a unified and consistent benchmark. Through systematic label harmonization and the inclusion of multiple textual representations, the dataset addresses a major gap in Bengali sentiment and hate speech research, particularly for data-hungry deep learning models. Baseline experimental results confirm that BengaliSent140 supports effective learning across a range of classical and neural architectures, establishing its value as a benchmarking resource.

Future work includes extensive evaluations using transformer-based models and large language models (LLMs), which are expected to better capture contextual and semantic nuances. The dataset can also be extended beyond binary labels into fine-grained hate categories such as political, religious, or personal hate using LLM-assisted annotation followed by human verification. Additionally, incorporating code-mixed and multilingual data would further enhance the dataset’s applicability to real-world Bengali digital communication.

\bibliographystyle{IEEEtran}
\bibliography{references}

\end{document}